\documentclass[runningheads]{llncs}
\usepackage{graphicx}
\usepackage[misc]{ifsym} 
\usepackage{balance} 
\usepackage{algpseudocode}
\usepackage{algorithm} 
\usepackage{stfloats}
\usepackage{appendix}
\usepackage{multirow} 
\usepackage{xcolor} 
\newcommand{\RNum}[1]{\uppercase\expandafter{\romannumeral #1\relax}}

\usepackage{enumitem} 
\usepackage{hyperref}
\begin{document}
\title{MTIRL: Multi-trainer interactive reinforcement learning system}
%
%
\author{Zhaori Guo$^{(\textrm{\Letter})}$ \and
Timothy J. Norman \and
Enrico H. Gerding}
\authorrunning{Zhaori Guo et al.}

\institute{University of Southampton, UK\\
\email{zg2n19@soton.ac.uk, t.j.norman@soton.ac.uk, eg@ecs.soton.ac.uk}}
\maketitle              
\begin{abstract}
Interactive reinforcement learning can effectively facilitate the agent training via human feedback. However, such methods often require the human teacher to know what is the correct action that the agent should take. In other words, if the human teacher is not always reliable, then it will not be consistently able to guide the agent through its training. In this paper, we propose a more effective interactive reinforcement learning system by introducing multiple trainers, namely Multi-Trainer Interactive Reinforcement Learning (MTIRL), which could aggregate the binary feedback from multiple non-perfect trainers into a more reliable reward for an agent training in a reward-sparse environment. In particular, our trainer feedback aggregation experiments show that our aggregation method has the best accuracy when compared with the majority voting, the weighted voting, and the Bayesian method. Finally, we conduct a grid-world experiment to show that the policy trained by the MTIRL with the review model is closer to the optimal policy than that without a review model.
\keywords{Interactive Reinforcement Learning \and Human-in-the-loop Reinforcement learning \and Multiple People Decision}
\end{abstract}
\section{Introduction}
\emph{Reinforcement Learning} (RL) is a machine learning method to train an agent to select actions in an environment to maximize a cumulative reward. 
In RL, the agent can usually be rewarded only at the end or at a particular state. 
This makes it difficult for the reward designer to quickly influence the knowledge of the agent in the key states.
Therefore, the reward sparsity has limited the application scenarios of reinforcement learning \cite{knox2008tamer,macglashan2017interactive}. 
One approach to solve this problem is the \emph{interactive RL} (IRL), which allows humans to participate in the training process.
Indeed, it has been shown that, through human feedback, the agent's learning is facilitated and, therefore, it is able to complete the training faster. In previous IRL research, people often focus on the interaction between a single human trainer and an agent \cite{knox2008tamer,macglashan2017interactive,knox2010combining}.  
However, this kind of methods require a perfect trainer, that is a trainer whose feedback is always correct. Indeed, it has been shown that if the quality of the trainer's feedback is not perfect, it cannot effectively help an agent finish training \cite{kurenkov2020ac}. For example, if the trainer gives wrong feedback on a critical state, the task will not be able to continue. Therefore, when a single trainer's trust is not enough, multiple trainers can make feedback more stable and reliable.

There are many scenarios in which MTIRL can be used fruitfully. 
Consider, for example, the asset portfolio management task, in which the agent has to decide on the best investment at every instant. Assuming that the trainer has a perfect knowledge of the market fluctuation is unfeasible. 
Another example is the patient's treatment. In this case, the experience of more doctors in training a medical agent is more reliable than a singular individual. 
To summarize, there are several contests in which one point of view is not enough to guarantee the level of knowledge that the agent needs to learn.
Due to the novelty of the approach, the IRL community has not been studying multi-trainers methods in full.
In this paper, we design the multi-trainer interactive reinforcement learning system (MTIRL), which can aggregate the binary feedback of multiple trainers into a reward that guides the agent training.
The logical structure of MTIRL is composed of four parts. 
The first part is the feedback setting. Since it is important to reduce the cognitive load on trainers as much as possible, we use binary feedback, that is, every trainer can express opinions through a good-or-bad question. 

The second part is the trust model, which takes care of understanding which of the trainers is more reliable than the others. We express the trustfulness of trainer $x$ through a parameter $P(x)$, which roughly estimates the probability that the report made by trainer $x$ is correct.
The third part is the decision model, which, given the feedback and trustworthiness of the trainers, allows the system to decide whether the agent should receive a positive or negative reward.
To decide what reward we should give to the agent, we combine a Bayesian model with a weighted voting model.
The last part is the reviewing one, which allows the system to remember all the agent's answers in a specific state-action pair. Thanks to the reviewing part, the model is able to correct previously unreliable feedback by updating the trust levels and by asking more and different trainers.

The main contributions of the paper are the following: we propose and study a new aggregating method, and we compare it with the already known ones through several experiments. The results show that MTIRL has better accuracy.
Moreover, the review model used in MTIRL has a smaller overall training cost and improves the accuracy. 
It is worthy of notice that the MTIRL method can be used as an alternative to all the single trainer IRL models.

The remainder of the paper is structured as follows. In Section \ref{Sec: rewo}, we provide an overview of related work. In Section \ref{Sec: modelform}, we formalize the problem. In Section \ref{Sec: MTIRL}, we introduce the details of the MTIRL system. The experimental setting is described in Section \ref{Sec: mtexper}. Finally, in Section \ref{Sec:conclu} we summarize the results of the paper and conclude.

\section{Related Work} \label{Sec: rewo}
To begin with, several IRL works focus on how to improve the RL performance by human feedback \cite{knox2008tamer,macglashan2017interactive}, or how to use implicit feedback to give reward \cite{cui2020empathic,bignold2021persistent}. These two methods only allow one trainer to participate in the RL training loop and cannot aggregate feedback from multiple trainers. 

In contrast, \cite{zhan2016theoretically,li2018optimal} enables multiple trainers to take part in IRL. However, they only select one trainer each time. If the quality of feedback from a single trainer is inaccurate, this will produce poor training accuracy. \cite{kurenkov2020ac} filters trainer's bad advice by comparing Q-values. However, this approach relies on environmental rewards as ground truth. It will not work well when the rewards are highly sparse. In addition, the feedback of multiple people cannot be effectively used, which will cause more costs. Our approach does not rely on ground truth, and it can effectively utilize all feedback.

In addition, there is also a lot of research about how to aggregate the feedback from multiple participants. One approach is to complete the aggregation by using the similarity between users \cite{goel2020personalized,zhong2022non}. This is usually implemented by similarity matrix and distance function. The purpose of them is to obtain a consensus that is in line with the majority's preferences. Therefore, the consensus does not have correct or wrong, but the reward in RL has. In contrast, our work allows the agent to judge the correct probability of the reward and how much trust should it give to the reward.

Another approach to aggregate feedback is using weighted voting. This approach assigns weights to participants in different ways and accumulates them as the result of the final aggregation \cite{tittaferrante2021multi,fan2020decentralized}. For example, \cite{tittaferrante2021multi} assigns fixed weights to different participants and aggregates them, whereas \cite{fan2020decentralized} uses the self-trust of participants to be the weights. The problem with the weighted voting method is that, if there is no ground truth, it cannot accurately represent the credibility of their aggregated results.

A third method is the distributed method, such as distributed RL and federated learning \cite{cao2021fltrust,ma2021federated}. These methods usually do not aggregate the information for training, but pass the training results from the local to the center. They are efficient, but only if the rewards obtained from the environment are reliable, this also needs the ground truth to judge the noise from local thread. Also, these methods require higher computing power. 

\section{Problem Formalization} \label{Sec: modelform}
In this paper, we solve a multiple trainers decision problem without any ground truth in IRL. In more detail, the RL model is represented by a Markov Decision Process (MDP). It consists of 5 elements ${\langle\mathcal{S}, \mathcal{A}, \mathcal{T}, R, \gamma\rangle}$, where: $\mathcal{S}$ is the set of states, also called the state space; $\mathcal{A}$ is the action space; $\mathcal{T}$ is the state transition function; and $R$ is the reward function. Furthemore, $\gamma \in [0,1]$ is the discount coefficient, which indicates the degree of influence of future rewards on the current state value. At each time $t$, the agent observes a state $s_t \in \mathcal{S}$. Then, it selects an action $a_t \in \mathcal{A}$ by policy $\pi(s_t|a_t)$; After this, it receives a reward $r_t$. The purpose of the RL task is to maximize the accumulated reward $R_t = \sum_{t=0}^{T} \gamma_t r_t$. In IRL, if human feedback is used as a reward, then it can be expressed as human reward $r'_t$.

In multi-trainers IRL, let $x \in X$ denote a set of trainers. We assume that more than half participants have a trust greater than the basic accuracy. After the agent performs action $a_t$, a set of trainers $Y_t \subseteq X$ gives feedback according to the state action pair $(s_t,a_t)$. The feedback set can be denoted as $F_t$. In IRL, if human feedback is used as a reward, it can be expressed as human reward $r'_t$. Trainers' rewards are different from environmental rewards, and they may not be correct. Therefore, the system needs to use $F_t$ to determine the final reward $r'_t = f(F_t)$ to improve the accuracy of the reward. Let $r^*_t$ denote the correct rewards. Our goal is to maximize the number $n$ of $r'_t = r^*_t$ under the feedback set $F_t$ by the function $f(F_t)$. It can express as:
\begin{equation}
	f^*(F_t) = \mathop{\arg\max}_{f(F_t)}n(r'_t = r^*_t|f(F_t)).
\end{equation}

\section{Multi-trainer Interactive Reinforcement Learning} \label{Sec: MTIRL}
\label{sec:mtirlformal}
\begin{figure}[ht]
   \centering
   \includegraphics[scale=0.38]{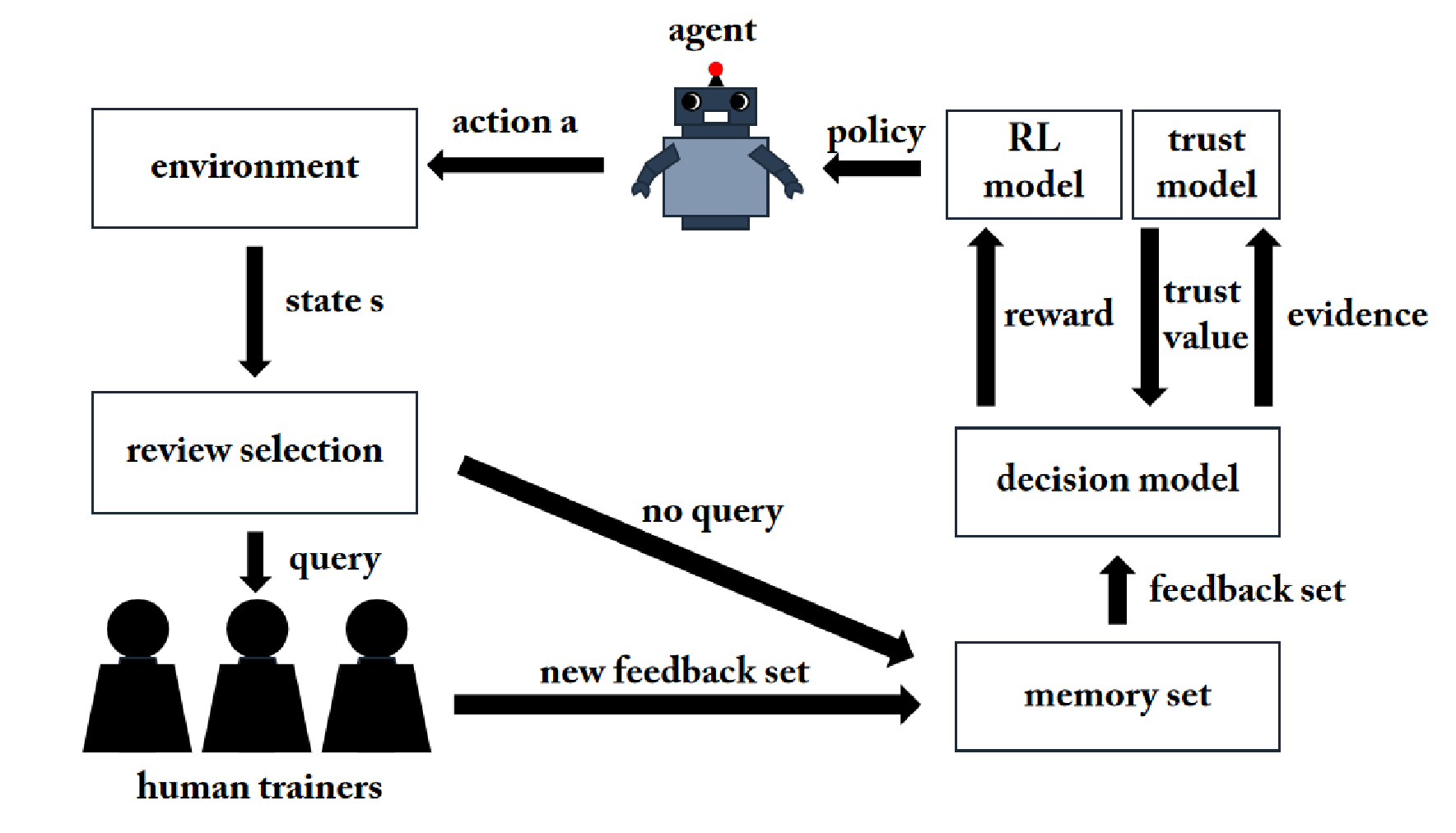}
   \caption{The feedback selection model decides whether to query trainers according to the state-action pair. The decision model combines the trust value and the feedback set to determine the final reward given to the agent. The RL model updates the policy by the reward. The decision model sends the evidence to the trust model for updating trainers' trust after the decision. }
   \label{Fig:MTIRLreview}
\end{figure}

MTIRL can aggregate the binary feedback of multiple trainers into an effective reward to help the agent training, which greatly improves the learning efficiency of the agent. The design of MTIRL considers four parts. The first part is the feedback setting. The second part is the trust model. The third part is the decision model. The last part is the review part. Figure \ref{Fig:MTIRLreview} shows its structure.

\subsection{Binary Feedback} \label{sec:feedsetting}
In order to reduce the feedback pressure on trainers, MTIRL uses binary reward feedback $f \in \{r_{pos}, r_{neg}\}$. If the human believes the action taken by the agent is the best, then a positive reward $r_{pos}$ is given. Otherwise, it is given a negative reward $r_{neg}$. There are two reasons for choosing it. Firstly, in the design of the feedback method, we plan to reduce the cognitive burden of human trainers as much as possible. Compared to the scoring reward, the binary reward can give humans less feedback pressure because humans only need to judge whether the current agent is performing the best action. Secondly, the binary reward is more robust to noise in feedback because it requires less cost than other methods to correct the noise.

\subsection{Trainer Trust Model} \label{sec:trustformal}
Modeling the trustworthiness of trainers is essential because everyone's knowledge level is different, so the quality of their feedback is also different. Therefore, if the system aggregates their feedback, it is necessary to understand their trustworthiness. 

\emph{subjective logic} is a method for modeling trust that is widely used \cite{burnett2010bootstrapping,gunecs2017budget,cheng2021general}. There are three reasons to use it. Firstly, it can quantify the trustworthiness of the participants and the uncertainty of the trust. It provides each trainer with a trust score between 0 and 1 to measure the reliability of the trainer. Secondly, it can identify malicious users to defend against attacks. Thirdly, it can be combined with Bayesian methods for decision-making.

\subsubsection{Evidence and Uncertainty} \label{sec:trustformaleau}
Given a set of trainers $x \in X$, let $b_x \in [0,1]$ be the agent belief to trainer $x$, and $d_x \in [0,1]$ denote the agent disbelief to trainer $x$. $u_x \in [0,1]$ is the uncertainty assessment. It decreases from 1 as evidence increases. For example, 300 trials have less uncertainty in a coin toss experiment than three trials. $a_x$ is a base rate , representing a prior degree of the trust before the model gets the evidence. If $a_x$ is set to 0.5, it means that the trainers randomly select positive or negative with equal probability. The relationship between $b_x$, $d_x$ and $u_x$ satisfies:
    \begin{equation}
    \label{eq:bdua}
        b_x + d_x + u_x = 1
    \end{equation}
 The trainer's trustworthiness $P{(x)}$ can be expressed as:
  \begin{equation}
  \label{Equ:P(x)}
    {P}{(x)}=b_{x}+a_{x} u_{x}
  \end{equation}
After each feedback, the results provide the trainer with positive evidence $\alpha_x$ or negative evidence $\beta_x$ observed by the agent, and then the system updates $b_x$, $d_x$ and $u_x$ as follows:
  \begin{equation}
  \label{eq:bdudetail}
    b_{x}=\frac{\alpha_x}{\alpha_x+\beta_x+2}, d_{x}=\frac{\beta_x}{\alpha_x+\beta_x+2}, u_{x}=\frac{2}{\alpha_x+\beta_x+2} 
  \end{equation}
The number 2 in the equation represents the weight of uncertainty \cite{josang2016subjective}. The methods for updating $\alpha_x$ and $\beta_x$ are introduced in Section \ref{sec:dmute}.

\subsection{Bayesian and Weighted Voting Ensemble Decision Model} \label{sec:bawvedm}
In this paper, we used a Bayesian and Weighted Voting Ensemble (BWVE) model to aggregate feedback from trainers. BWVE will calculate the correct probability of positive and negative reward under the feedback set from trainers, respectively. After that, the system will choose the one who has the bigger correct probability as the aggregated reward. There are two reasons for choosing BWVE. Firstly, it can effectively utilize the trustworthiness of trainers to make the aggregation results more accurate. Secondly, its results are probability, which can measure the reliability of the aggregated reward. 

\subsubsection{Bayesian Probability} \label{sec:bawvedmbp}
In RL, in each time $t$, the agent selects an action $a_t$ under state $s_t$. A set of trainers $P_{t} \subseteq X$ gives positive feedback, whereas set of trainers $N_{t} \subseteq X$ gives negative feedback, and $P_{t} \cap N_{t} = \emptyset$. Under the condition $P_{t}$ and $N_{t}$, the probability of the positive or negative feedback is right can be denoted as $P_t(f_{pos}|P_{t}, N_{t})$ and $P_t(f_{neg}|P_{t}, N_{t})$, respectively. Through the Bayesian method, they can be expressed as:
  \begin{equation}
  \label{Equ:P(fpc)}
    P_t(f_{pos}|P_{t}, N_{t}) = \frac{P(f_{pos})P_t(P_{t}, N_{t}|f_{pos})}{P(f_{pos})P_t(P_{t}, N_{t}|f_{pos})+P(f_{neg})P_t(P_{t}, N_{t}|f_{neg})}
  \end{equation}
  \begin{equation}
  \label{Equ:P(fnc)}
    P_t(f_{neg}|P_{t}, N_{t}) = \frac{P(f_{neg})P_t(P_{t}, N_{t}|f_{neg})}{P(f_{pos})P_t(P_{t}, N_{t}|f_{pos})+P(f_{neg})P_t(P_{t}, N_{t}|f_{neg})}
  \end{equation}
The $P(f_{pos})$ and $P(f_{neg})$ are the basic probability. For example, in the absence of prior information, if the trainer can randomly give the agent a $r_{neg}$ or $r_{pos}$ reward with the same probability, $P(f_{pos})$ and $P(f_{neg})$ are 0.5. $P_t(P_{t}, N_{t}|f_{pos})$ or $P_t(P_{t}, N_{t}|f_{neg})$ means given the condition that positive or negative feedback is correct, the probability that decision is correct under the condition $P_{t}$ and $N_{t}$. They can be computed using the trustworthiness $P(x)$ mentioned in section \ref{sec:trustformal}. So it can be expressed as:
  \begin{equation}
  \label{Equ:P(cfp)}
    P_t(P_{t}, N_{t}|f_{pos}) =\prod_{i \in P_t} \prod_{j \in N_t} P(i)(1-P(j))
  \end{equation}
    \begin{equation}
  \label{Equ:P(cfn)} 
    P_t(P_{t}, N_{t}|f_{neg}) =\prod_{i \in P_t} \prod_{j \in N_t}(1-P(i))P(j)
  \end{equation}
\subsubsection{Weighted Voting Trust Initialization} \label{sec:bawvedmwvti}
In the initialization phase of the trust model, since the trust probability $P(x)$ of each trainer is not reliable, the performance of the Bayesian decision-making method is not reliable, which means that the Bayesian method are vulnerable to bad feedback. We use the weighted voting method to solve this problem because it is more stable in the initialization phase. The trainer's trustworthiness $P(x)$ is considered weights in the weighted voting method. The difference from the Bayesian method is that it only adds up each trainer's trust weight. The correct probability of positive and negative through weighted voting can be expressed as:
  \begin{equation}
  \label{Equ:P(fpcv)}
  P^{wv}_t(f_{pos}|P_{t}, N_{t}) = \frac{\sum_{i \in P_t}P(i)}{\sum_{j \in P_t \cup N_t}P(j)}
  \end{equation}
   \begin{equation}
  \label{Equ:P(fncv)}
  P^{wv}_t(f_{neg}|P_{t}, N_{t}) = \frac{\sum_{i \in P_t}P(i)}{\sum_{j \in P_t \cup N_t}P(j)}
  \end{equation}
\subsubsection{Bayesian and Weighted Voting Ensemble Decision Method}
After getting the Bayesian (Equation \ref{Equ:P(fpc)} and  \ref{Equ:P(fnc)}) and weighted voting probability (Equation \ref{Equ:P(fpcv)} and  \ref{Equ:P(fncv)}), the system uses the average uncertainty $\bar{u}_{t}$ of the trainers to combine and balance the Bayesian and the voting method. It represents the confidence or initialization level of the trust model. It can be expressed as:
   \begin{equation}
  \label{Equ:ut}
  \bar{u}_{t}=\frac{\sum_{i \in P_t \cup N_t}u_{i}}{n(P_t \cup N_t)}
  \end{equation}
The average uncertainty $\bar{u}_{t}$ decreases as time $t$ increases, which means that the trust model becomes more and more believable. The correct probability of positive and negative through ensemble method can be expressed as:
  \begin{equation}
  \label{Equ:P(fpfinal)}
   P^{ag}_t(f_{pos}|P_t, N_t) = (1-\bar{u}_{t})P_t(f_{pos}|P_t, N_t) + \bar{u}_{t}P^{wv}_t(f_{pos}|P_t, N_t)
  \end{equation}
   \begin{equation}
  \label{Equ:P(fnfinal)}
   \quad \quad P^{ag}_t(f_{neg}|P_t, N_t) =(1-\bar{u}_{t})P_t(f_{neg}|P_t, N_t) + \bar{u}_{t}P^{wv}_t(f_{neg}|P_t, N_t)
  \end{equation}
As $\bar{u}_{t}$ decreases, the weights of the weighted voting methods are decreasing and the weights of the Bayesian methods are increasing. $P^{ag}_t(f_{pos}|P_t, N_t)+P^{ag}_t(f_{neg}|P_t, N_t)$ = 1.

\subsubsection{Decision Making and Update Trust Evidence} \label{sec:dmute}
After aggregating the feedback, the system needs to compare $P^{ag}_t(f_{pos}|P_t, N_t)$ and $P^{ag}_t(f_{neg}|P_t, N_t)$. If $P^{ag}_t(f_{pos}|P_t, N_t)$ is bigger than $P^{ag}_t(f_{neg}|P_t, N_t)$, action $a_t$ is considered to be the best action, the aggregating reward $r'_t$ = $r_{pos}$. Otherwise, action $a$ is not the best action, the aggregating reward $r'_t$ = $r_{neg}$.

In Section \ref{sec:trustformal}, the computation of trainers' trust probability is closely related to evidence $\alpha$ and $\beta$. But the problem is that there is no ground truth to allow the agent to judge whether the reward $r'_t$ is reliable in reward sparsity tasks. The difference of $P^{ag}_t(f_{pos}|P_t, N_t)$ and $P^{ag}_t(f_{neg}|P_t, N_t)$ can be the feedback confidence value $i_t$:

  \begin{equation}
  \label{Equ:P(itcon)}
 i_t =  |P^{ag}_t(f_{pos}|P_t, N_t)-P^{ag}_t(f_{neg}|P_t, N_t)|
  \end{equation}
Here, $i_t$ is a measure of the confidence of the aggregated reward. The greater $i_t$, the higher the credibility of $r'_t$. Evidence $\alpha_x$ and $\beta_x$ updates should also be affected by $i_t$. Therefore, we use $i_t$ as the step size for the evidence update. 
If $P^{ag}_t(f_{pos}|P_t, N_t) > P^{ag}_t(f_{neg}|P_t, N_t)$, it means the positive answer is more likely to be correct, so the evidence $\alpha_{x}$ and $\beta_{x}$ is updated for each trainer using the following equation:
   \begin{equation}
  \label{Equ:vtqt}
\begin{array}{l}
    \forall x \in P_{t},\quad \alpha_{x}\leftarrow \alpha_{x}+i_{t} \\\\
    \forall x \in N_{t},\quad \beta_{x}\leftarrow\beta_{x}+i_{t}
\end{array}
  \end{equation}
Similarly, if $P^{ag}_t(f_{neg}|P_t, N_t) > P^{ag}_t(f_{pos}|P_t, N_t)$,  the negative answer is correct, and so we update $\alpha_{x}$ and $\beta_{x}$ using:
   \begin{equation}
  \label{Equ:vtqt}
\begin{array}{l}
    \forall x \in P_{t},\quad \beta_{x}\leftarrow\beta_{x}+i_{t} \\\\
    \forall x \in N_{t},\quad \alpha_{x}\leftarrow\alpha_{x}+i_{t}
\end{array}
  \end{equation}
\subsection{Feedback Review Model} \label{mtirlreviewm}
Human feedback is often expensive and time-consuming. In an IRL system, the designer needs to reduce the pressure of human feedback as much as possible. One way to reduce feedback is to avoid giving rewards to the state-action pair that has been feedback before. So in the MTIRL system, the system records feedback from trainers in each state-action pair. If the agent reencounters the same state, it uses the reward in the memory, greatly reducing the number of human feedback. However, the problem with doing this is that the future rewards are wrong if the previous reward is wrong. A review mechanism is applied in the MTIRL system to correct the wrong reward. The system re-asks those state-action pairs which have unreliable rewards.

$P_{s,a}$ and $N_{s,a}$ denote the positive and negative historical feedback sets of the state-action pair $(s, a)$. At the time $t$, if the agent encounters a state-action pair that has already been asked, it needs to decide whether to ask the trainers again. The system needs to calculate $P^{ag}_t(f_{pos}|P_{s,a}, N_{s,a})$ (equation \ref{Equ:P(fpfinal)}) and $P^{ag}_t(f_{neg}|P_{s,a}, N_{s,a})$ (equation \ref{Equ:P(fnfinal)}) again with the latest trust model and $P_{s,a}$ and $N_{s,a}$. Then, the system can get the new reward $r'_{s,a}$ and the confidence value $i'_{s,a}$. The probability of review $P_{re}(s_t, a_t) \in [0,1]$ can define as:
  \begin{equation}
  \label{Equ:Pf(stat)}
    P_{re}(s_t, a_t) = 1 - i'_{s,a}
  \end{equation}
If $i'_{s, a}$ is larger, it means that the information of historical feedback is more reliable, and the probability of review is lower. On the contrary, if $i'_{s,a}$ is smaller, it means that the information of historical feedback is unreliable, and the probability of review is higher. If it is decided not to review, then the system can use the new reward $r'_{s,a}$ as reward $r'_t$. If it decides to query again, then the system firstly needs to combine the historical feedback with the new feedback to form a new feedback set $P'_t \leftarrow P_{s,a} \cup P_t$, $N'_t \leftarrow N_{s,a} \cup N_t$, and secondly calculate $P^{ag}_t(f_{pos}|P'_t, N'_t), P^{ag}_t(f_{neg}|P'_t, N'_t)$ again to make the decision.

\section{Experiments} \label{Sec: mtexper}
In this paper, we designed two experiments. The first is a trainer feedback aggregation experiment, which is used to verify the performance of the trust model and the decision model. The second is grid-world experiment, which is used to test the performance of the review model and MTIRL. Our hypotheses are as follows:
\begin{enumerate}[label=H1]
\item The BWVE is more accurate in estimating the correct reward in aggregating feedback from multiple trainers with varying trustworthiness than the Bayesian estimation, weighted voting and simple majority voting.
 \label{hyp: bwve}
\end{enumerate}
\begin{enumerate}[label=H2]
\item The use of confidence measures calculated by BWVE enables MTIRL to construct policies from trainers that are close to the optimal policy.
 \label{hyp: MTIRL}
\end{enumerate}
In order to verify the reliability of the experimental results, we used Mann–Whitney U test combined with Bonferroni Correction to test the difference between the results of every two methods. 

\subsection{Trainer Feedback Aggregation} \label{sec:qat}
In this experiment, we test Hypotheses \ref{hyp: bwve} to evaluate our BWVE model (Section \ref{sec:bawvedm}). We do this through an agent questioning multiple trainers and using their feedback to estimate the feedback that would be received from a truthful advisor. (i.e. the correct reward)
They are the Bayesian method (Equation \ref{Equ:P(fpc)} and \ref{Equ:P(fnc)}), weighted voting method (Equation \ref{Equ:P(fpcv)} and \ref{Equ:P(fncv)}), majority voting method (weighted voting method without trust model,every trainer's trust is 1), BWVE (Equation \ref{Equ:P(fpfinal)} and \ref{Equ:P(fnfinal)}). 

\subsubsection{Trainer Feedback Aggregation Experimental Setting}
There are 1000 independent questions with binary answers. Fifty simulated trainers participated in the experiment. An agent needs to answer 1000 questions in sequence, and it will query trainers for help before answering each question. Trainers do not always answer questions after query, whereas each trainer has a $10\%$ probability to give feedback to each question. 

Moreover, different trainers have different real trustworthiness. For example, if a trainer's real trustworthiness is 80$\%$, he has 80$\%$ probability to give a correct answer. The trainers' real trustworthiness is generated by extended rectified Gaussian distribution with different mean and standard deviation \cite{palmer2017methods}.  To simulate the different trust distributions, we designed 6 groups of experiments with different standard deviations of trainers' trustworthiness (0, 0.1, 0.2, 0.3, 0.4, 0.5). The different standard deviations influence the difference between trainers' trust. Each group of experiments generated trainers' trust from a mean of 0.51 to 1 (increase 0.01 each set, a total of 50 sets), respectively. 

After the agent has answered 1000 questions, we will count the number of correctly answered questions and calculate the accuracy under different aggregating methods as results. Each set of experiments was run 100 times. Table \ref{tab:qaset} shows the trainer feedback aggregation experimental setting.

\begin{table}[ht]
\centering
\caption{In every different method, there are 1000 questions and 50 trainers. 10$\%$ trainers give their answer in every question. Every set of experiment runs 100 times. There are 6 groups of experiments with different standard deviations (0, 0.1, 0.2, 0.3, 0.4, 0.5). Each group of experiments generated trainers' trust from a mean of 0.51 to 1.}\label{tab:qaset}
\begin{tabular}{p{21em}|p{12em}}
\hline
{Description} & Value\\

\hline
    {Total number of questions} & 1000\\
    {Total number of trainers} & 50  \\
    {Trustworthiness mean of trainers} & From 0.51 to 1  \\
    {Trustworthiness standard deviations of trainers} & 0, 0.1, 0.2, 0.3, 0.4, 0.5  \\
    {Feedback probablity} & 10\%  \\
    {Experimental times} & 100  \\
\hline
\end{tabular}
\end{table}

\subsubsection{Trainer Feedback Aggregation Experimental Results} \label{sec:qatr0}
\begin{figure*}[ht]
   \centering
   \includegraphics[scale=0.2]{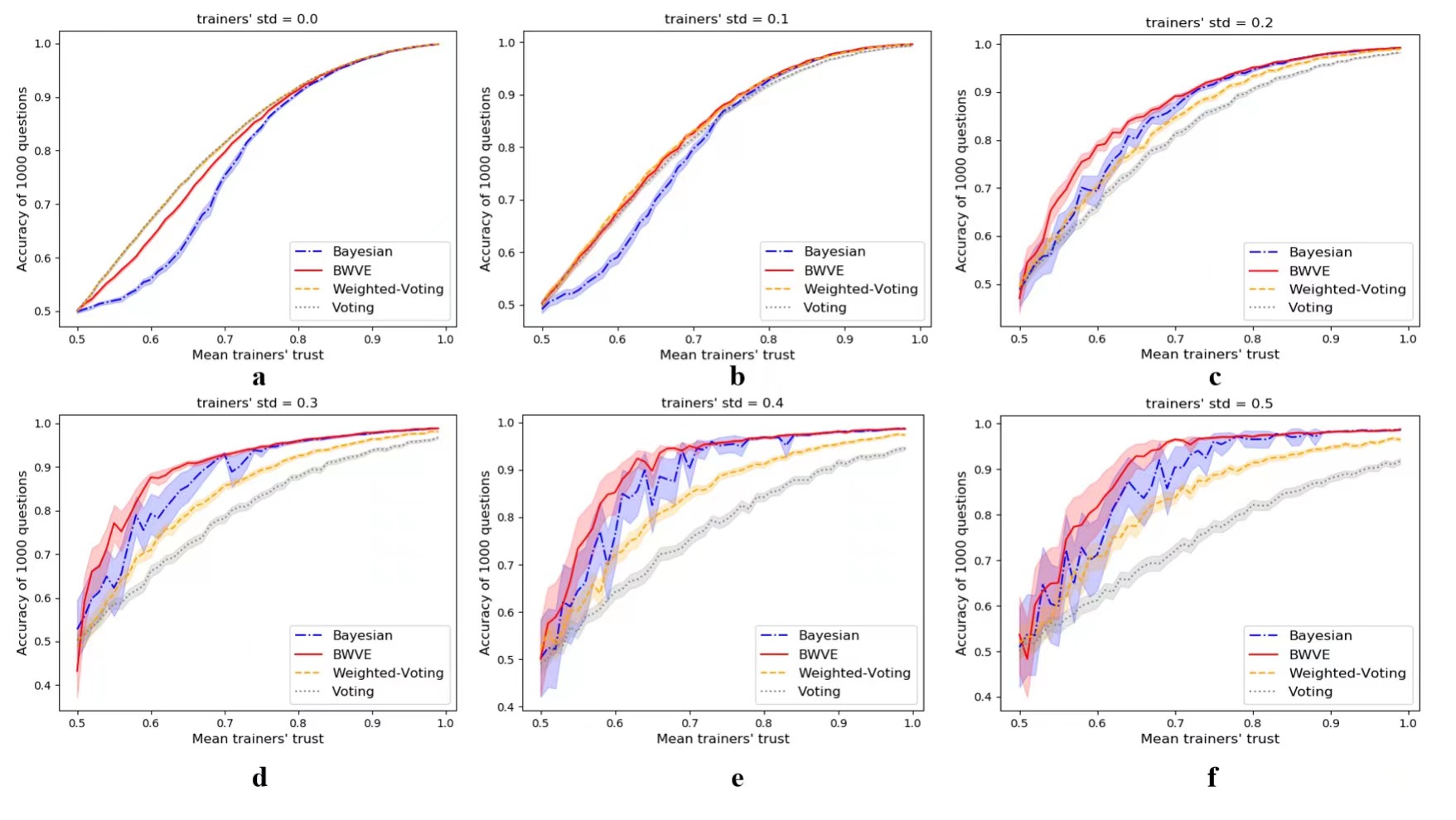}
   \caption{This results are the answer accuracy as the trust mean increases (under 0 and 0.5 standard deviations of trainers' trust). The X-axis represents the mean of trainers' trust from 0.51 to 1. The Y-axis represents the average answer accuracy of 100 experiments. There are 50 points in every curve. Each point is the mean of 100 experimental results. The part with transparent color is the 95$\%$ confidence interval error bar.}
   \label{Fig:qar012345}
\end{figure*}

Figure \ref{Fig:qar012345} shows the six group of results of trainer feedback aggregation experiments. 
Firstly, in varying standard deviation and mean of trainers' trustworthiness, BWVE almost has the best accuracy against Bayesian, weighted voting, and majority voting. Only if the trustworthiness of trainers is smaller than 0.75 and all trainers have identical trustworthiness (i.e. Figure \ref{Fig:qar012345}a, the standard deviation of trainers' trustworthiness is 0), the voting methods is a little better than BWVE. We would argue that 0 standard deviation of trainers' trustworthiness in an MTIRL scenario is unlikely. In 900 pairs of the Mann-Whitney Test about BWVE, there are 487 (more than 50$\%$) pairs $p<0.0083 (0.05/6)$. This supports hypotheses \ref{hyp: bwve}. 

Secondly, compared to the Bayesian method in 6 sets of results, the error bar of BWVE is smaller than that of the Bayesian method in all the results, so BWVE is more stable than the Bayesian method. 

Thirdly, the higher the standard deviation of trainers' trust, the more advantages Bayesian and BWVE methods over voting methods. When the standard deviation is greater than 0.2 (Figure \ref{Fig:qar012345}c, d, e, f), The average accuracy of BWVE and Bayesian method is better than the voting method.

Fourthly, the higher the standard deviation of the trainers' trust, the more efficient the trust model. Except for the result of 0 standard deviations (Figure \ref{Fig:qar012345}a) where the weighted voting and voting methods perform similarly, the rest show that the weighted voting consistently outperforms the voting method. In most results, the three other methods involving trust models are also preferred over voting methods.

Overall, BWVE considers the voting methods' stability and the Bayesian method's high accuracy to achieve better performance. In the initialization phase, the trust model is not reliable, so the Bayesian method doesn't work. Whereas BWVE can rely on the voting part to initialize trainers' trust, and then it gradually biases the decision weights towards the Bayesian part as the trust uncertainty decreases. So it almost always performs better than voting methods and has more stability than Bayesian methods.

\subsection{MTIRL} \label{sec:MTIRLtest}
In this experiment, we test Hypotheses \ref{hyp: MTIRL} to evaluate our review model (Section \ref{mtirlreviewm}) by grid-world task. We set up two additional multi-trainer IRL methods to compare with our MTIRL system. The first is MTIRL-no review. It does not use the review model and only provides feedback once for each state-action pair. The previous reward is used for the same state-action pair. The second is MTIRL-unlimited, which is that the agent gets feedback after every taking action.

\subsubsection{Gird-world Experimental Setting}
\begin{figure*}[ht]
   \centering
   \includegraphics[scale=0.2]{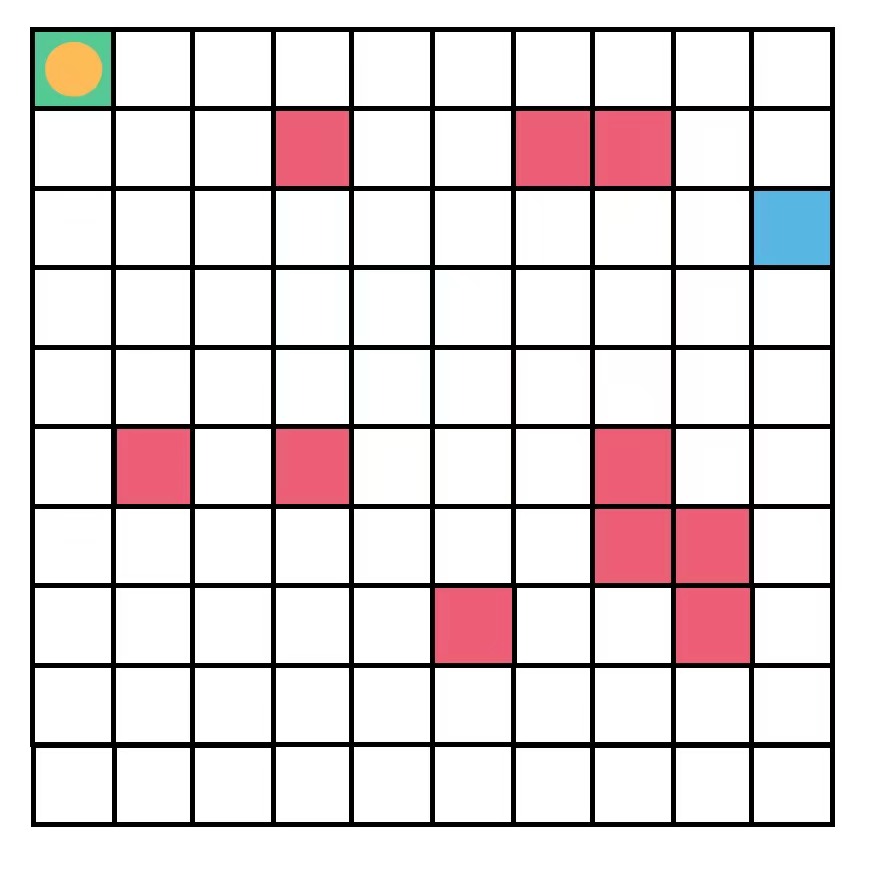}
   \caption{There are 100 states in total.The green gird is the start state; the blue grid is the goal state; the white grid is the normal states. and the red grid represents the cliff states.}
   \label{Fig:gridworld}
\end{figure*}

The experiment uses 10*10 grid-world as the experimental environment. It is a classic environment for testing the performance of RL algorithms and has been used in many studies \cite{knox2015framing,kazantzidis2022train}. Its module consists of 6 kinds of parts: agent, start state, goal state, normal state, and cliff state (Figure \ref{Fig:gridworld}). The start state (randomly selected in each episode) is where the agent is at the beginning of each game episode. If the agent reaches the goal state, it passes the game successfully. Normal states can walk freely, and the start state is also a normal state. If the agent reaches the cliff state, it will die. 

Table \ref{tab:mtirlset} shows the experimental setting. There are five trainers and the trustworthiness of trainers is set the same as that in the trainer feedback aggregation experiment, but their standard deviation of trustworthiness is 0.2. Each set of experiments also needs to be repeated 100 times. The maximum episodes of the 10*10 cliff grid-world are set to 500. This is because the IRL method completes training in around 150 episodes without noise interference. Taking into account the effect of noise, we scaled up the maximum episodes. Moreover, the agent is easily going into a loop because of the noise, so the agent can perform at most 200 actions in each episode for saving computational power costs. If the agent finds the best solutions, it also stops training and records the results.

In the experiments, we use the state–action–reward–state–action (SARSA) algorithm as the basic algorithm for the experiment \cite{rummery1994line}. In interactive SARSA, its environmental rewards are replaced by human rewards. A convergent Q-table trained by the SARSA algorithm was used to simulate humans to give reward in the experiments. 
\begin{table}[ht]
\centering
\caption{The max episodes are 500. Max number of actions in every episode is 200. Every experiment runs 100 times. There are 5 trainers.}\label{tab:mtirlset}
\begin{tabular}{p{21em}|p{12em}}
\hline
{Description} & Value\\
\hline
    {Max episodes} & 500\\
    {Max action} & 200  \\
    {Number of Trainers} & 5  \\
    {Trustworthiness means of trainers} & From 0.51 to 1  \\
    {Trustworthiness standard deviations of trainers} & 0.2  \\
    {Feedback probablity} & 100\%  \\
    {Experimental times} & 100  \\
\hline
\end{tabular}
\end{table}

\subsubsection{MTIRL Girdworld Performance Results} \label{sec:MTIRLtestpr}
Firstly, the policy trained by the MTIRL with the review model is closer to the optimal policy than that without a review model. At a mean of trainers' trustworthiness from 0.55 to 0.9, MTIRL with the review model has better performance than MTIRL-unlimit and MTIRL-no review methods. Three MTIRL methods have similar performance when the mean is less than 0.55 and greater than 0.9. In 100 Mann-Whitney hypotheses test results between MTIRL and the others without review model, 38 has $p< 0.0166 (0.05/3)$, so it supports Hypotheses \ref{hyp: MTIRL}. Although the MTIRL-unlimit method can provide infinite feedback, indiscriminate feedback may revise the correct feedback into a wrong one. So that its feedback maintains a fixed accuracy rate, its performance is not prominent. This also shows that the review feedback without some evidence is not working to improve the efficiency of agent learning.

\begin{figure*}[ht]
   \centering
   \includegraphics[scale=0.12]{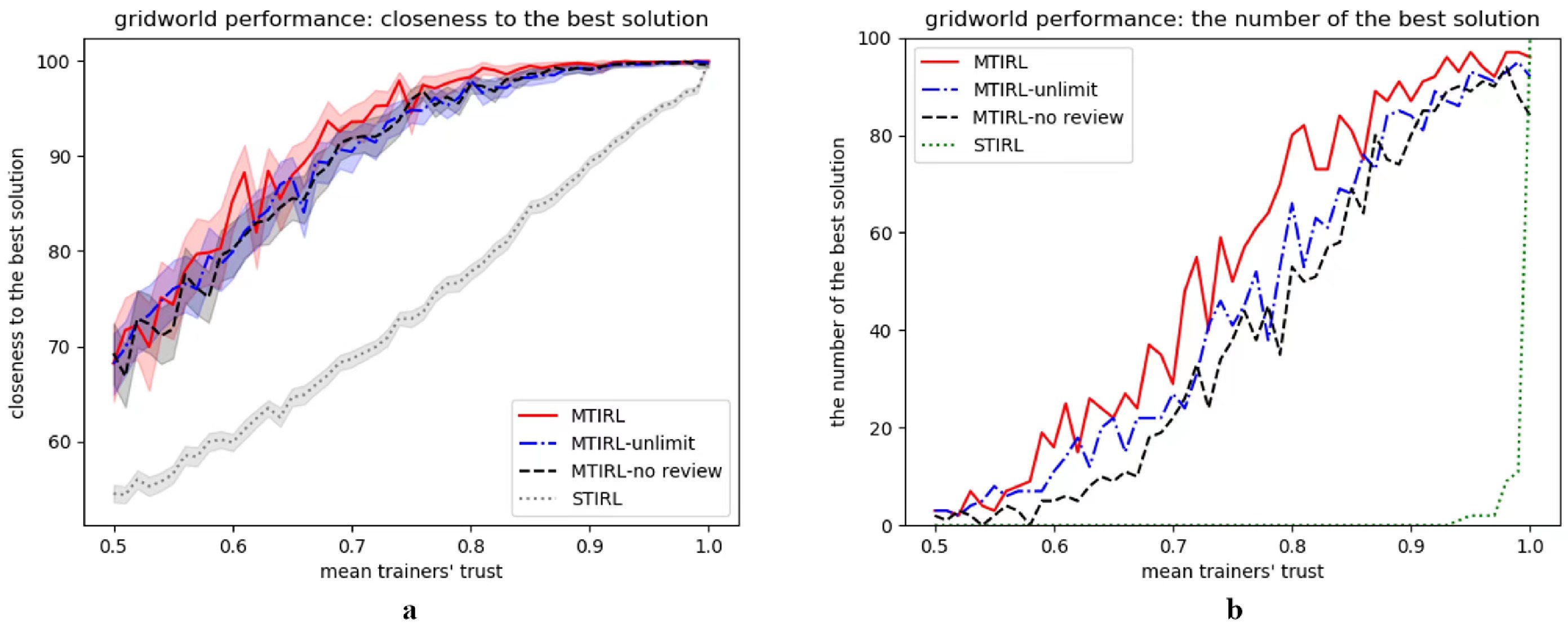}
   \caption{The figure a shows the closeness to the best solution as the trainers' trust means increases. The X-axis represents the mean of trainers' trust from 0.51 to 1. The Y-axis represents the average closeness to the best solution at the end of training in 100 experiments. There are 50 points in every curve. Each point is the mean of 100 experimental results. The part with transparent color is the 95$\%$ confidence interval error bar. Figure b shows the number of the best solution in 100 experiments, as the trainers' trust means increases. The X-axis represents the mean of trainers' trust from 0.51 to 1. The Y-axis represents the number of the best solution in 100 experiments. There are 50 points in every curve.}
   \label{Fig:mtirlasbs}
\end{figure*}

\begin{figure*}[ht]
   \centering
   \includegraphics[scale=0.12]{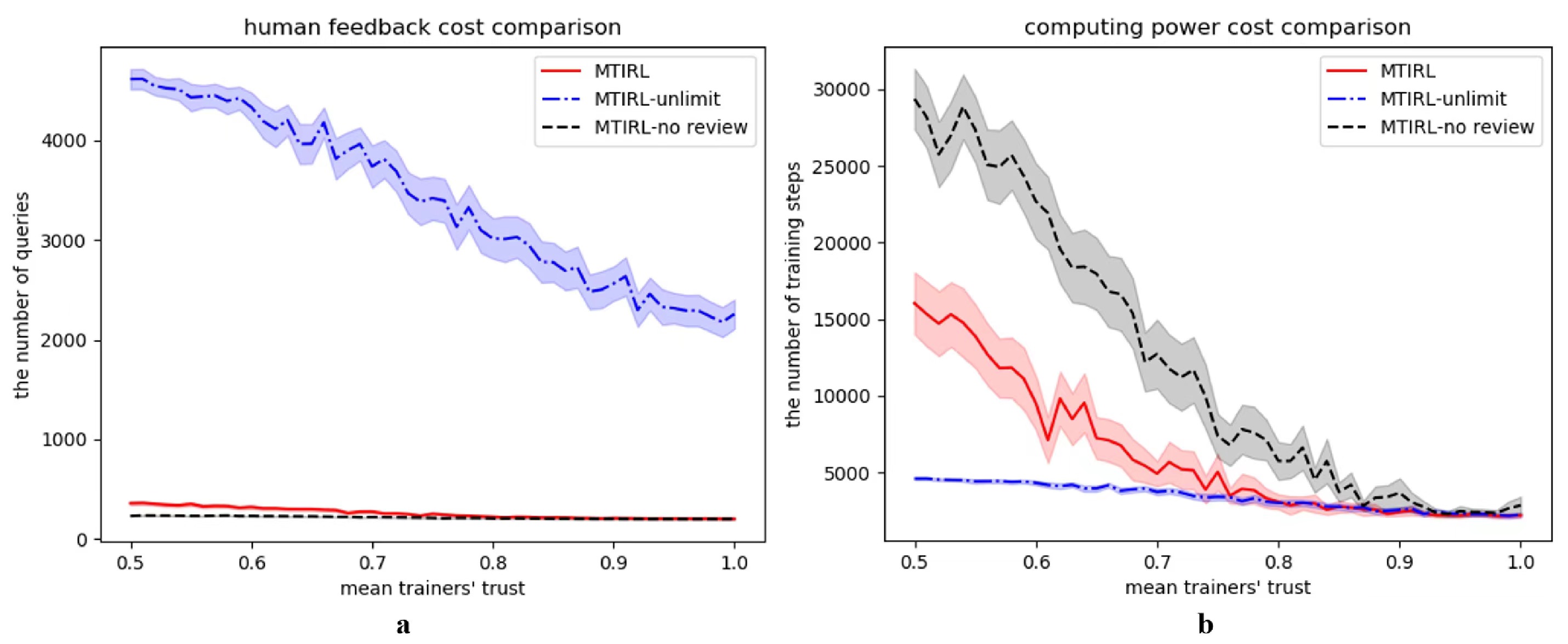}
   \caption{Figure a shows the average number of queries in 100 experiments as the trainers' trust means increases. The X-axis represents the mean of trainers' trust from 0.51 to 1. The Y-axis represents the average number of queries in 100 experiments. There are 50 points in every curve. Each point is the mean of 100 experimental results. The transparent color area is the 95$\%$ confidence interval error bar. Figure b shows the average number of training steps in 100 experiments as the trainers' trust means increases. Different from Figure a, The Y-axis represents the average number of training steps in 100 experiments.}
   \label{Fig:mtirlhc}
\end{figure*}

Secondly, the review model uses a small amount of feedback to improve the learning ability of the agent. In Figure \ref{Fig:mtirlhc}a, MTIRL-unlimit has the most feedback times and is around ten times more than the other two methods. MTIRL and MTIRL-no review have relatively low feedback costs, and MTIRL is slightly higher than MTIRL - no review. As can be seen from the figure, the amount of feedback from MTIRL decreases as the mean trainers' trust increases. When the trainers' trust is relatively low, the confidence value is very low, making the probability of review very large. 

Thirdly, the review model reduces the risk of trapping the agent in a loop. Figure \ref{Fig:mtirlhc}b shows the average number of training steps of three different MTIRL methods in 100 experiments under different trust means. MTIRL-no review uses the most steps. When the mean is less than 0.8, the average training steps of MTIRL-no review are almost twice that of MTIRL. This means that if we only trust the first feedback result of each state-action pair, then the agent easily falls into a loop because of the wrong feedback. The MTIRL-unlimit method uses the fewest steps but requires a huge human cost. This is because, if the previous answer puts the agent in the loop, then the infinite queries have a big chance to change their answer, which can make the agent break out of the loop. Moreover, for MTIRL, because of the review model, there is also a chance that it will ask the trainers again, thus changing the wrong answer and getting the agent out of the loop.

Finally, compared to single trainer IRL, MTIRL has a more powerful performance. In Figure \ref{Fig:mtirlasbs}a, experimental results show that when the average trust of trainers is 0.7, the agent can learn the best solutions of $90\%$, while the single-trainer IRL can only learn around $70\%$. Before the mean trainer's trust was $95\%$, it was almost difficult for the single-trainer IRL method to learn all the best solutions, but MTIRL can learn the best solutions many times. 

\section{Conclusion} \label{Sec:conclu}
In this paper, we proposed MTIRL, which is the first IRL system that can combine advice from multiple trainers. It can aggregate a set of feedback from non-perfect trainers into a more reliable reward for RL agent training in a reward-sparse environment. We use a question-answer experiment to support the Hypotheses \ref{hyp: bwve}, that is, the BWVE is more accurate in estimating the correct reward in aggregating feedback from multiple trainers with varying trustworthiness than the Bayesian estimation, weighted voting, and simple majority voting. We also use the grid-world experiment to test Hypotheses \ref{hyp: MTIRL}. The results show that the policy trained by the MTIRL with the review model is closer to the optimal policy than that without a review model. 

In future work, we will work on the optimization of two parts. First, we will consider the limit on the number of feedback trainers can give, that is, the system needs to maximize the value of each feedback, leveraging human feedback on more critical states. Second, the cost of each query needs to be considered, which means that the system should balance the cost and value to maximize the utility.
%
%
\bibliographystyle{splncs04}
\bibliography{MTIRL}
\end{document}